\documentclass[10pt,twocolumn,letterpaper]{article}

\usepackage{cvpr}              % To produce the CAMERA-READY version
\usepackage{dblfloatfix}
\usepackage{afterpage}
\definecolor{cvprblue}{rgb}{0.21,0.49,0.74}
\usepackage[pagebackref,breaklinks,colorlinks,allcolors=cvprblue]{hyperref}

\def\paperID{8107} % *** Enter the Paper ID here
\def\confName{CVPR}
\def\confYear{2026}

\usepackage{url}        
\usepackage{booktabs}       
\usepackage{xcolor}       
\usepackage{colortbl}
\usepackage{graphicx}
\usepackage{multirow} 
\usepackage{algorithmic}
\usepackage{multicol}
\usepackage{threeparttable}
\usepackage{bm}

\usepackage{cuted}
\usepackage{capt-of}
\usepackage{pifont}

\definecolor{ourblue}{rgb}{0.847,0.847,1}
\definecolor{ourpink}{rgb}{1,0.8,0.8}

\def\ourdata{SurgClean}
\newcommand{\tabref}[1]{Table \ref{#1}}

\newcommand{\cmark}{\ding{51}}
\newcommand{\xmark}{\ding{55}}%

\newcommand{\figref}[1]{Fig. \ref{#1}}

\newcommand{\secref}[1]{Sec. \ref{#1}}

\graphicspath{{./Imgs/}}
\DeclareGraphicsExtensions{.jpg,.pdf,.png}

\title{Benchmarking Endoscopic Surgical Image Restoration and Beyond}

\author{
Jialun Pei$^{1}$ \quad
Diandian Guo$^{1}$ \quad
Donghui Yang$^{2}$ \quad
Zhixi Li$^{3}$ \quad \\
Yuxin Feng$^{4}$ \quad
Long Ma$^{2}$\thanks{Corresponding author. (malone94319@gmail.com)} \quad
Bo Du$^{5}$ \quad
Pheng-Ann Heng$^{1}$ \\
[2mm]
$^1$ The Chinese University of Hong Kong \quad
$^2$ Dalian University of Technology \quad \\
$^3$ Southern Medical University \quad
$^4$ Xidian University \quad 
$^5$ Wuhan University \\
}

\begin{document}
\maketitle
\begin{strip}
	\vspace{-0.5em}
	\centering
    \includegraphics[width=0.96\linewidth]{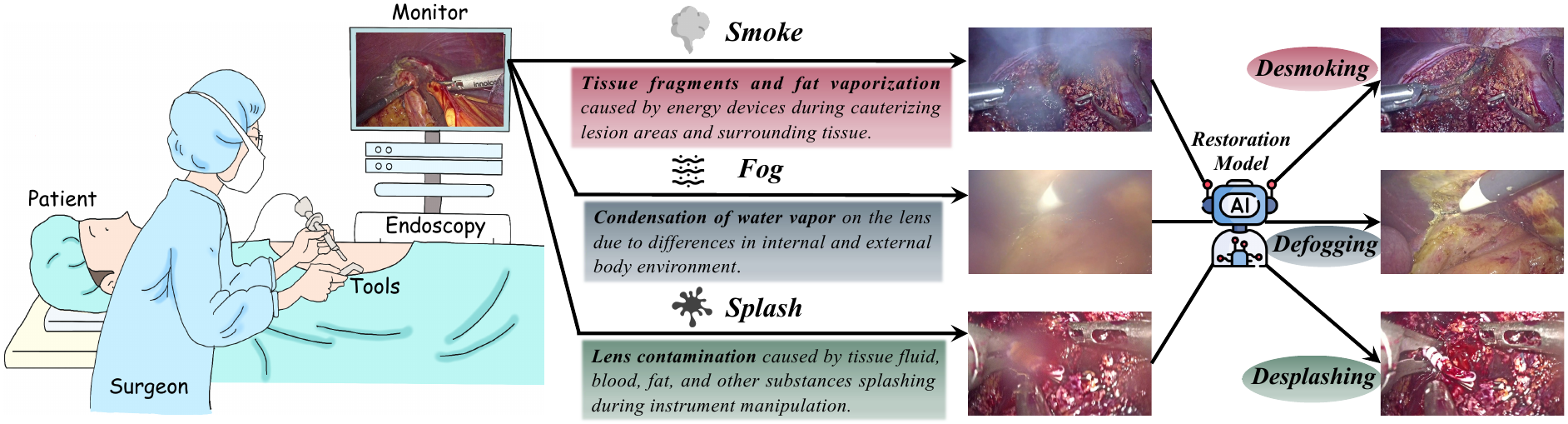}
	% \vspace{-0.3cm}
	\captionof{figure}{Illustration of surgical image restoration in endoscopic surgery, including endoscopic operating scenes, multi-type degradation causes and images, and algorithms for surgical restoration. }
	\label{fig:fig1}
    \vspace{-5pt}
\end{strip}

\begin{abstract}
In endoscopic surgery, a clear and high-quality visual field is critical for surgeons to make accurate intraoperative decisions.
However, persistent visual degradation, including smoke generated by energy devices, lens fogging from thermal gradients, and lens contamination due to blood or tissue fluid splashes during surgical procedures, severely impairs visual clarity.
These degenerations can seriously hinder surgical workflow and pose risks to patient safety.
To systematically investigate and address various forms of surgical scene degradation, we introduce a real-world open-source surgical image restoration dataset covering endoscopic environments, called~\textbf{\ourdata}, which involves multi-type image restoration tasks from two medical sites, \ie, desmoking, defogging, and desplashing. \ourdata~comprises 3,113 images with diverse degradation types and corresponding paired reference labels.
Based on~\ourdata, we establish a standardized evaluation benchmark and provide performance for 22 representative generic task-specific image restoration approaches, including 12 generic and 10 task-specific image restoration approaches.
Experimental results reveal substantial performance gaps relative to clinical requirements, highlighting a critical opportunity for algorithm advancements in intelligent surgical restoration.
Furthermore, we explore the degradation discrepancies between surgical and natural scenes from structural perception and semantic understanding perspectives, providing fundamental insights for domain-specific image restoration research.
% Our work aims to empower the capabilities of restoration algorithms to clean surgical scenes and improve the efficiency of clinical procedures.
Our work aims to empower restoration algorithms and improve the efficiency of clinical procedures.
% Data and code are available. 
Sources can be available at: \href{https://github.com/PJLallen/Surgical-Image-Restoration}{https://github.com/PJLallen/Surgical-Image-Restoration}.
\end{abstract}    
\vspace{-14pt}
\section{Introduction}
\textbf{Backgrounds}: 
% Image restoration techniques, such as dehazing, de-raining, and deblurring, have achieved significant advancements in natural scenes~\citep{dai2022flare7k,feng2024advancing,xiao2022image}.
% However, their exploration and application in clinical medicine, particularly in minimally invasive surgery, remains relatively limited. In the restricted space of endoscopy, a clear surgical field of view and improved image quality in the operative area are critical for preventing unexpected complications and ensuring surgical safety~\citep{alapatt2024jumpstarting,gruter2024surgical,pei2025synergistic}. In actual endoscopic surgery, visual clarity could be severely impaired by various intraoperative factors, \eg, smoke generated by energy instruments, fog formation on lenses due to temperature differences, and lens contamination caused by splashing of blood or tissue fluids during surgical manipulations, as illustrated in~\figref{fig1}. 
In minimally invasive surgery, a clear surgical field of view and improved image quality in the operative area are critical for preventing unexpected complications and ensuring surgical safety~\citep{hong2023mars,li2025efficient}.
Especially in narrow endoscopic environments, visual clarity could be severely impaired by various intraoperative factors, \eg, smoke generated by energy instruments, fog formation on lenses due to temperature differences, and lens contamination caused by splashes of blood or tissue fluids during surgical manipulations, as shown in~\figref{fig:fig1}. 
Diverse types of degradation seriously hinder surgeons' observation and judgment of surgical conditions, making it difficult to identify anatomy and lesion sites, potentially leading to incorrect operations and reducing surgical success rates~\citep{pei2025instrument}. 
Thus, endoscopic surgical image restoration (ESIR) is critical to providing surgeons with clear and stable intraoperative views, significantly reducing surgery time, minimizing surgical errors, and improving operational efficiency.

\noindent\textbf{Challenges}:
Unlike natural scenarios, the endoscopic environment poses unique challenges for image restoration, characterized by diverse types and levels of image degradation, which seriously distort the surgical field of view and significantly hinder the distinction of anatomical structures~\citep{guo2025surgical,pei2024depth,zhou2025landmark}.
As illustrated in~\figref{fig:fig1}, \emph{Smoke} originates from numerous fine tissue fragments and fat vaporization when surgeons use energy devices to cauterize diseased areas and surrounding tissues. 
% The smoke mainly diffuses irregularly outward from the operating area. 
\emph{Fog} is mainly caused by condensation of water vapor when the temperature gradient between the intra- and extra-corporeal environments leads to mist forming on the lens.
Compared to smoke noise, the distribution of surgical fog is broader and more uniform.
For \emph{Splashing}, the use of instruments such as electrocautery and ultrasonic scalpels during surgery results in the splashing of blood, tissue fluid, and other substances onto the lens, causing contamination and increasing the difficulty of restoration.
Moreover, the restoration process requires cleaning anatomical details, \eg, blood vessels and tissue structures, to assist doctors in making precise surgical decisions. 
% Although endoscopic equipment has achieved relatively high clarity, image distortion, color discrepancies, and saturation imbalance may still occur. 
When complex degradations are superimposed, critical clues obscured by noisy substances may blend into the background, further increasing degradation complexity.
% To this end, SIR presents unique complexities and challenges, highlighting the urgency of developing dedicated benchmark datasets tailored to surgical restoration tasks.
To this end, ESIR presents unique complexities and challenges, highlighting the urgency of developing dedicated benchmark datasets for surgical restoration.

\noindent\textbf{Datasets}:
Existing ESIR datasets usually have limitations in terms of authenticity and diversity. 
% Several surgical restoration datasets~\citep{sengar2021multi,chen2019smokegcn,wang2023surgical} are generated through manual synthesis, resulting in discrepancies with real-world data and restricting the applicability of algorithms in actual scenarios. 
Several surgical restoration datasets~\citep{sengar2021multi,chen2019smokegcn,wang2023surgical} are generated through manual synthesis, restricting the applicability of algorithms in actual scenarios. 
% Other datasets derived from real surgical scenes, \ie, Cyclic-DesmokeGAN~\citep{venkatesh2020unsupervised}, Desmoke-LAP~\citep{pan2022desmoke}, and LSVD~\citep{wu2025self}, mainly focus on a single type of degradation, such as desmoking. 
Other datasets derived from real surgical scenes~\citep{venkatesh2020unsupervised,pan2022desmoke,wu2025self} mainly focus on a single type of degradation, such as desmoking. 
To comprehensively investigate degradation challenges in real-world endoscopic surgery, we construct the first multi-source ESIR dataset, \emph{\ourdata}, aimed at effectively assisting surgeons in reducing visual noise in the operating field of view. 
% \ourdata~comprises 3,113 endoscopic images under various degradation conditions selected from real surgical videos of 414 patients, accompanied by adjacent clean frames as paired reference labels. 
\ourdata~comprises 3,113 endoscopic images under various degradations selected from real surgical videos of 414 patients across two medical sites, accompanied by adjacent clean frames as paired reference labels.
Driven by clinical needs, our dataset introduces three challenging ESIR tasks: \emph{desmoking}, \emph{defogging}, and \emph{desplashing}, and further subdivides the degradation levels according to difficulty and characteristics.
This dataset enables researchers to systematically explore and evaluate various image restoration methods under real endoscopic conditions, providing opportunities for broader research and algorithm development.

\noindent\textbf{Benchmarks}:
% With the launch of~\ourdata, we establish a comprehensive SIR benchmark to address the inherent data, theoretical, and computational challenges in this field.
We establish a comprehensive ESIR benchmark to address the inherent data, properties, and computational challenges based on~\ourdata.
For desmoking, defogging, and desplashing, we evaluate 22 representative image restoration methods, including general and task-specific frameworks.
Extensive experimental results reveal that: 
% 1) Although existing image restoration techniques perform effectively in natural scenes, their accuracy and efficiency are insufficient to meet clinical requirements when dealing with complex multi-factor degradation, especially desplashing.
1) Although existing image restoration techniques perform effectively in natural scenes, their accuracy and efficiency are insufficient to meet clinical requirements when dealing with complex multi-factor degradation.
2) While current methods show reasonable performance for samples with lower-level degradation, they fail to produce noticeable improvements for severe degradation at higher levels.
% 3) The restored results have limited benefits for downstream surgical understanding tasks and may even have a negative impact.
% 3) The restored results have limited benefits for downstream surgical understanding tasks.
These findings highlight the significant potential and necessity for developing advanced algorithms specifically designed for ESIR.
By contributing to this benchmark community, we aim to guide future research prospects and methodological innovations, ultimately enhancing the efficiency of minimally invasive surgery and reducing surgical risks.

Our main contributions are three-fold: 
\textbf{(1)} We conducted a thorough exploration of various types of degradation in real-world surgical scenarios and introduced~\ourdata, the first multi-source ESIR dataset in endoscopic surgery.
\textbf{(2)} We established a comprehensive ESIR benchmark based on~\ourdata, achieving a performance evaluation of multiple types and levels of restoration methods.
\textbf{(3)} We revisited various degradation characteristics specific to surgical conditions and performed a manifold discrepancy analysis with degradation in natural scenes, intending to explore potential optimization strategies for surgical degradation.

\section{Related Works}

\begin{figure*}[t]
\centering
\includegraphics[width=0.98\linewidth]{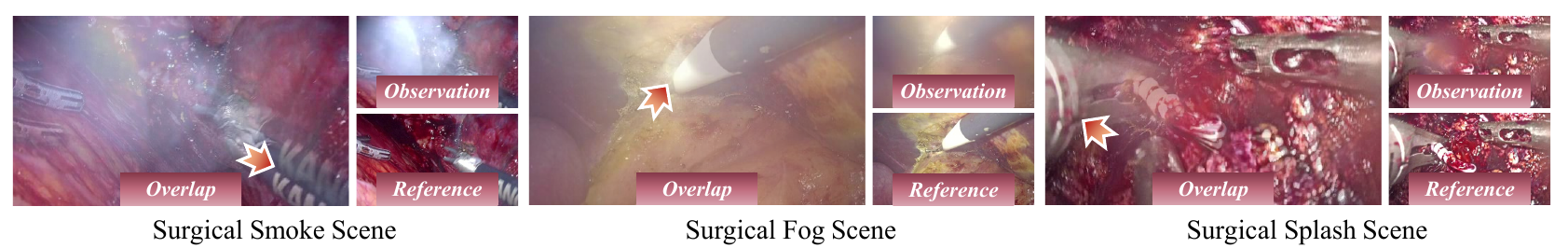}
\vspace{-0.6em}
\caption{Illustration of degraded images and corresponding unaligned reference labels in~\ourdata. The red arrows indicate noticeable ghosting artifacts in the overlapping regions.}
\label{data_gt}
\vspace{-10pt}
\end{figure*}

\begin{figure}[t!]
\centering
\includegraphics[width=\linewidth]{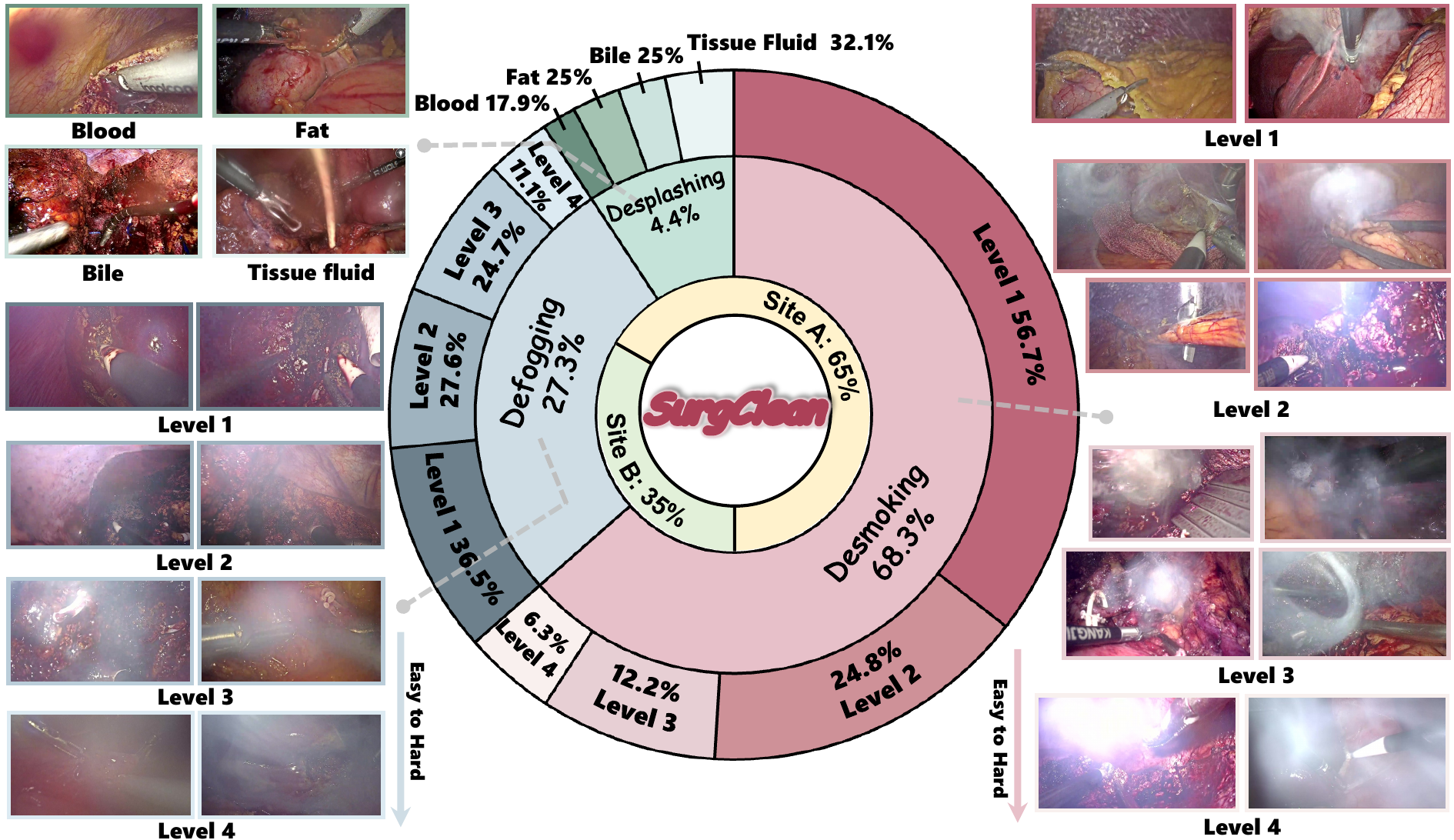}
\vspace{-12pt}
\caption{Illustration of the proposed~\ourdata~dataset. We exhibit data distribution under different degradation scenarios and further provide fine-grained divisions according to interference levels and attributes. Typical samples are displayed on both sides.}
\label{dataset}
\vspace{-5pt}
\end{figure}

% \subsection{Surgical Restoration Datasets}
\textbf{Surgical Restoration Datasets.} 
Numerous surgical image restoration (SIR) datasets have been introduced for training and evaluation in surgical scenes~\citep{sengar2021multi,chen2019smokegcn,wang2023surgical,venkatesh2020unsupervised,pan2022desmoke,xia2024new}.
Many are synthetic post-processed sets~\citep{sengar2021multi,chen2019smokegcn,chang2024lsd3k,wang2023surgical}, creating a gap to clinical degradations; for example, SmokeGCN~\citep{chen2019smokegcn} overlays Gaussian-filtered smoke on clean images. These datasets enabled early progress but generalize poorly to complex, variable clinical conditions.
More recent real-surgery datasets emphasize desmoking, including Cyclic-DesmokeGAN~\citep{venkatesh2020unsupervised}, DeSmoke-LAP~\citep{pan2022desmoke}, DesmokeData~\citep{xia2024new}, and LSVD~\citep{wu2025self}; DeSmoke-LAP targets hysterectomy videos with unpaired translation, and LSVD mines smoke scenes using pre-smoke frames as references.
Despite their value, these resources typically address a single degradation and offer limited diversity. To bridge this gap, we introduce a comprehensive dataset covering multiple real-world surgical degradations.

\noindent\textbf{Image Restoration Approaches.} 
General image restoration methods broadly fall into three families: CNN-based, Transformer-based, and Mamba-based architectures. CNN models (e.g., ConvIR~\citep{cui2024revitalizing}, FocalNet~\citep{focal2023}) leverage multi-scale feature extraction for robust restoration. Transformer variants (Restormer~\citep{zamir2022restormer}, X-Restormer~\citep{chen2024xrestormer}, AST~\citep{zhou2024AST}, RAMiT~\citep{choi2024reciprocal}, Histoformer~\citep{sun2025restoring}) enhance modeling capacity via attention. Recent state-space models, notably Mamba, have motivated Mamba-based restorers such as MambaIR~\citep{guo2025mambair} and MambaIRv2~\citep{guo2024mambairv2}.
For defogging, methods include RIDCP~\citep{wu2023ridcp} with an enhanced haze model and codebook design, C$^2$PNet~\citep{zheng2023curricular} with contrastive learning, DEA~\citep{chen2024dea} with stronger feature extractors, CORUN~\citep{fang2024real} with collaborative unfolding, and CoA~\citep{ma2025coa} with a compress-then-adapt pipeline for synthetic-to-real transfer. For desmoking, DCP~\citep{he2010single} remains a simple baseline; MS-CNN~\citep{wang2019multiscale} combines Laplacian pyramids with CNNs; IIT-EDC~\citep{salazar2020desmoking} targets real-time GAN-based removal; Desmoke-LAP~\citep{pan2022desmoke} builds on CycleGAN; SelfSVD~\citep{wu2025self} implements self-supervision by utilizing clean frames from smoke-contaminated videos to guide the desmoking process. However, a unified restoration framework that robustly handles multiple surgical degradations remains largely unexplored.

\section{SurgClean Dataset}\label{data}

% Stemming from clinical requirements, our~\ourdata~dataset provides three challenging endoscopic surgical image restoration (ESIR) tasks in real-world clinical scenarios, \ie, desmoking, defogging, and desplashing.
% We provide a detailed overview of~\ourdata~in terms of data collection, data annotation, and statistics analysis.

\begin{table*}[t]
\centering
\scriptsize
\renewcommand{\arraystretch}{1.3}
\renewcommand{\tabcolsep}{1.7mm}
\caption{Comparing SurgClean with existing real-world surgical image restoration datasets. “OA” denotes public availability.}
\vspace{-1em}
\begin{tabular}{l|c|c|c|c|c|c|c|c|c}
% \hline
\toprule
 Dataset     & OA  &  Reference & Resolution & Desmoke & Defog & Desplash & Scale & Site No.    & Data Source        \\ \hline
Cyclic-DesmokeGAN~\citep{venkatesh2020unsupervised}  &   \xmark &   \xmark  &  240$\times$320      & \cmark&  \xmark&\xmark  & 1,400  & 1 & Cholec80  \\
Desmoke-LAP~\citep{pan2022desmoke}  & \cmark  & \xmark  &  720$\times$540      & \cmark & \xmark & \xmark  & 3,000  &  1 &Hysterectomy  \\ 
DesmokeData~\citep{xia2024new}  & \cmark   &  \cmark   &  700$\times$350      & \cmark& \xmark &  \xmark  & 961  & 1 &  da Vinci Si videos   \\ \hline
\textbf{\ourdata~(Ours)}  & \cmark   &  \cmark   &  \textbf{1280$\times$720}      & \cmark& \cmark & \cmark & \textbf{3,113}  & \textbf{2} & {\textbf{Gall, Bile, Liver, Pan, Spl; Lung, Med, Eso}} 
\\ 
% \hline
\bottomrule
\end{tabular}
\begin{tablenotes}
    \scriptsize \item Site A: \{Gall: Gallbladder, Bile: Bile Duct, Pan: Pancreatectomy, Spl: Splenectomy, Liver\}; Site B: \{Med: Mediastinal, Eso: Esophageal, Lung\}.
\end{tablenotes}
\vspace{-5pt}
\label{Data_Comp}
\end{table*}

\begin{table}[t]
\centering
\scriptsize
\renewcommand\arraystretch{1.2}
\renewcommand{\tabcolsep}{2.2mm}
\caption{Degradation level description in~\ourdata.}
\vspace{-1em}
\begin{tabular}{c|p{0.78\linewidth}}
\toprule
Level & Detailed Descriptions \\
\hline
\textbf{Level 1} & \textbf{Mild degradation:} Less than 1/3 of the entire field of view is filled with smoke or fog, and the anatomical structures in the operating area are visible and do not affect judgment. \\
\hline
\textbf{Level 2} & \textbf{Moderate degradation:} 1/3--2/3 of the entire field of view is filled with smoke or fog, and the anatomical structures in the operating area are still visible but less clearly than in Level 1, without affecting judgment. \\
\hline
\textbf{Level 3} & \textbf{Severe degradation:} More than 2/3 of the entire field of view is filled with smoke or fog, or the operating area is degraded, which may affect judgment. \\
\hline
\textbf{Level 4} & \textbf{Complete degradation:} More than 2/3 of the entire field of view is filled with smoke or fog, and the operating area is degraded, severely affecting judgment. Surgery must be paused to wait for the degradation to clear. \\
\bottomrule
\end{tabular}
\label{class-level}
\vspace{-10pt}
\end{table}

\subsection{Data Collection}

The degradation of surgical scenes caused by contaminants during endoscopy occurs according to the specific surgical procedure and has a certain degree of randomness. Therefore, we focus on restoring noise images that appear at specific time points rather than video-level restoration to improve clinical application efficiency. 
Concretely, we collect endoscopic surgical videos of 414 patients with a total duration of approximately 43,640 minutes undergoing laparoscopic and thoracoscopic surgeries from two medical sites.
Four intern surgeons are invited to pick out the frames from each video where they identified the surgical field as obscured by visual interference, including smoke, fog, and splash contamination on the lens. After that, two professional surgeons review and filter all interfering images to ensure reliability.
We gather a total of 3,113 noisy images with a resolution of 1280$\times$720, including 2,127 images for desmoking, 849 images for defogging, and 137 images for desplashing.
The data in each sub-task is divided into training and test sets in an 8:2 ratio.
\figref{dataset}~illustrates the data distribution of~\ourdata~across three types of ESIR from two medical sites.
The proportion of samples taken up by different sub-tasks also reflects the real-world occurrence rates of interference events during surgical procedures, \eg, the probability of splash is the lowest among three types.
\figref{dataset}~also exhibits various complex examples of visual interference encountered in endoscopic surgery.

\subsection{Data Annotations}\label{dataanalysis}
Since all samples in~\ourdata~are derived from real-world endoscopic surgical videos, it is difficult to obtain perfectly aligned ground truth frames. 
To provide relatively clear paired reference labels for deep learning-based model supervision, we adopt the PS-frame scheme in line with~\citep{wu2025self}, \ie, using the last clean frame preceding the noisy frame as the paired reference.
As shown in~\figref{data_gt}, this form of pairing may be affected by the movement of the endoscopic lens, resulting in slight deviations in pixel distribution. This also poses a key challenge for model supervision in this benchmark.
To reduce the impact of misaligned references on the restoration model during training, we uniformly employ the optical flow alignment method~\citep{zhang2021learning,sun2018pwc} to estimate the optical flow between the input image and the unaligned reference for relative displacement.
Then, we warp the reference using the estimated optical flow to transform the unaligned images into aligned pairs.
This strategy allows us to train and evaluate models in as same as general image restoration tasks for fairness and effectiveness.

\subsection{Data Statistics and Analysis}

\begin{figure}[t]
\centering
\includegraphics[width=\linewidth]{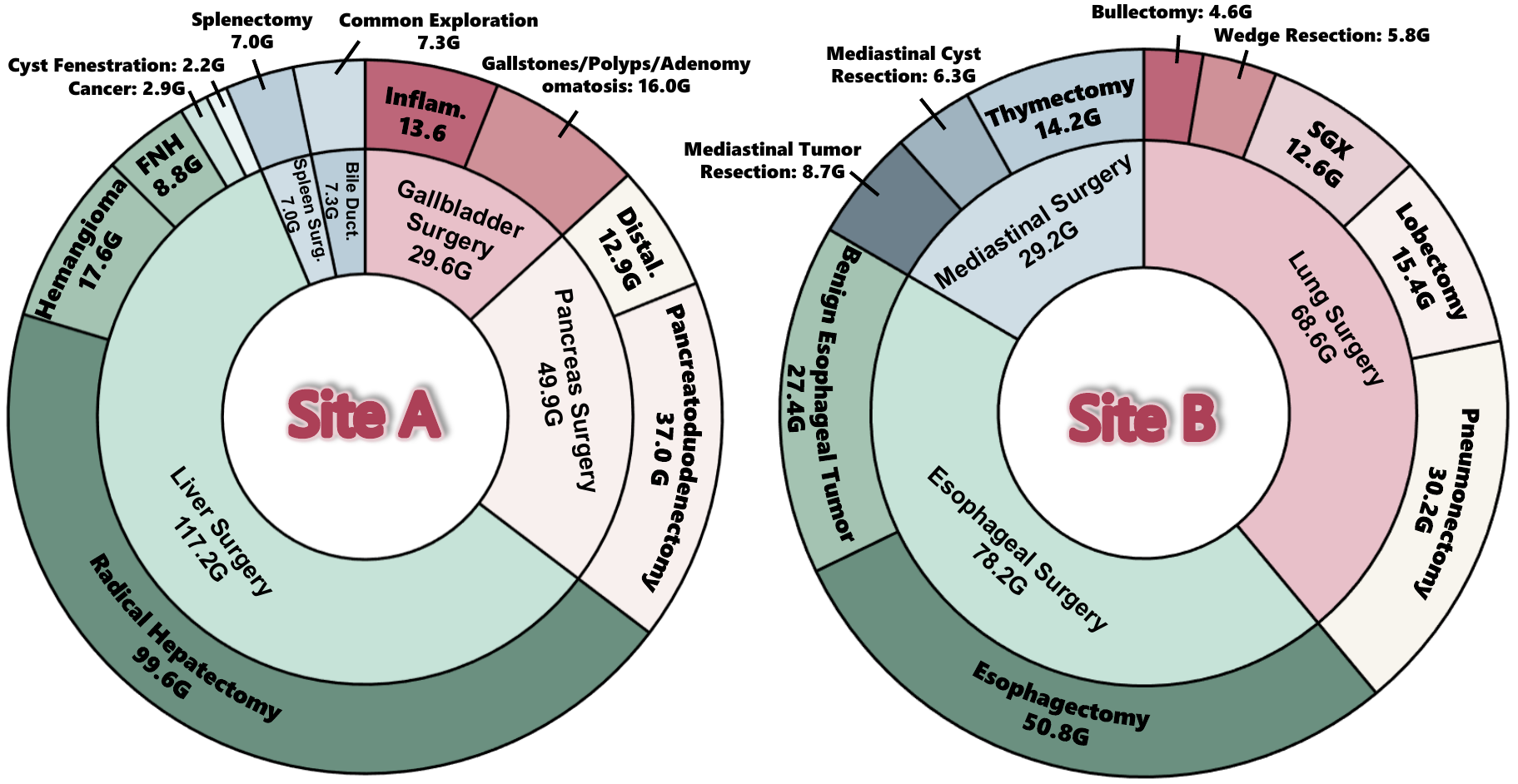}
\vspace{-1.0em}
\caption{Data statistics of \ourdata~from two medical sites, categorized by surgical procedure and detailed by surgical types.}
\label{video_size}
\vspace{-10pt}
\end{figure}

To facilitate fine-grained research, we further divided samples into multiple subsets based on different degradation levels and attributes. Specifically, for desmoking and defogging tasks, surgeons divide corresponding samples into four levels (from easy to difficult, \ie, Level 1-4) based on the area and degree of smoke or fog obstruction and the degree of interference with the operating zone. For desplashing, the samples are grouped into four subsets (\ie, $T_{blood}$, $T_{fat}$, $T_{bile}$, and $T_{fluid}$) based on the type of substances splashed onto the lens, including blood, fat, bile, and tissue fluid. 
We elaborate on the definition and classification of different levels of degradation in~\tabref{class-level}. In addition, \figref{dataset} exhibits typical examples of different difficulty levels and each type of splash interference. 
\figref{video_size}~also illustrates the data statistics of \ourdata~from two medical sites, 
showing the distribution of video sizes across surgical procedure categories and further subdivisions by specific procedure types.
More sample exhibitions and statistics of~\ourdata~dataset can be found in the supplementary material.

Furthermore, we summarize and perform comparisons with other real-world ESIR datasets. As illustrated in~\tabref{Data_Comp}, existing ESIR datasets focus solely on desmoking, while~\ourdata~comprehensively embraces multiple types of image degradation in real-world surgical procedures. 
In addition, datasets with paired reference labels provide more significant support for model fine-tuning.
Compared to other datasets with a single data source, our data source comes from various types of endoscopic surgery, \eg, gallbladder, pancreatectomy, and esophageal, to accommodate data diversity.
\ourdata~comprises a variety of interference surgical scenarios to promote comprehensive evaluation.
% 此外，具有配对参考标签可以促进基于深度学习复原模型的训练和评估。我们在补充材料中也与包含配对标签的DesmokeData数据集进行了交叉验证实验，以验证我们数据集的有效性和挑战性。
Additionally, the paired reference labels facilitate the training and evaluation of deep restoration models. 
% We conduct cross-validation experiments with DesmokeData~\citep{xia2024new} to validate the effectiveness and challenging nature of our dataset (see \pjl{\emph{Supp. E}}).
% 同时，我们也在Supp. C中将我们数据集的降质图像与其他ESIR数据集进行了可视化对比。
% We also perform a visual comparison of degraded images in our dataset with other datasets in \emph{Supp. C}.

% Our dataset has passed the ethics statement and can be publicly available at \url{https://kaggle.com/datasets/5bad41858571a3a9ea2f65a50c1d1d81c71956cc966c5b6ab96a42fa46418d78}.

\section{Benchmarking and Results}\label{benchmark}

% We provide an overview of benchmark construction and evaluation for surgical image restoration (SIR) based on~\ourdata~dataset. The benchmark includes implementation details, evaluation metrics, training setting, and performance evaluation and analysis of various representative image restoration models, including generic and task-specific image restoration models.

% All pre-trained weights and evaluation codes of our benchmark are available at \url{https://github.com/PJLallen/Surgical-Image-Restoration}.

\subsection{Experimental Setup}

\textbf{Implementation Details.}
% (480 images for training, 120 images for testing)
% (256 images for training, 64 images for testing)
% (80 images for training, 20 images for testing)
All methods are implemented with PyTorch~\citep{paszke2017automatic} and trained on two NVIDIA RTX 4090 GPUs with the Adam optimizer. During training, the input image is randomly cropped into 128$\times$128 patches with a batch size of 2 and a total of 200,000 iterations. In addition, we use random crop and horizontal flips as data augmentation. The initial learning rate of all methods is set to default value, and the learning rate is decreased by half every 100k iterations throughout the training process.
Additionally, all models are trained on the training set of each sub-scenario and then individually inferred on the corresponding test set on~\ourdata.
For desmoking, 1,700 images are employed for training and 427 for testing. For defogging, 679 images are used for training and 170 for testing. 
% For desplashing, the subset divides the training and test samples in an 8:2 ratio.
The subset of desplashing divides the training and test sets in an 8:2 ratio.

\begin{table*}[t]
    \centering
    \scriptsize
    \renewcommand\arraystretch{1.1}
    \renewcommand{\tabcolsep}{0.7mm}
    \caption{Quantitative comparison for surgical image restoration with representative generic restoration models on the~\ourdata~test sets. Best and second-best among all methods are highlighted in \textcolor{pink}{pink} and \textcolor{blue}{blue}, respectively.}
    \vspace{-1em}
    \begin{tabular}{l|l|ccccc|ccccc|ccccc|r}
        \toprule
        \multirow{2}{*}{\textbf{Methods}} & \multirow{2}{*}{\textbf{Pubs}} &
        \multicolumn{5}{c|}{\textbf{Desmoking}} &
        \multicolumn{5}{c|}{\textbf{Defogging}} &
        \multicolumn{5}{c|}{\textbf{Desplashing}} &
        \multirow{2}{*}{\textbf{Params}} \\
        \cline{3-17}
        & & PSNR$\uparrow$ & SSIM$\uparrow$ & LPIPS$\downarrow$ & NIQE$\downarrow$ & PI$\downarrow$
          & PSNR$\uparrow$ & SSIM$\uparrow$ & LPIPS$\downarrow$ & NIQE$\downarrow$ & PI$\downarrow$
          & PSNR$\uparrow$ & SSIM$\uparrow$ & LPIPS$\downarrow$ & NIQE$\downarrow$ & PI$\downarrow$
          & \\
        \hline
        Restormer~\citep{zamir2022restormer} & CVPR'22 & 18.939 & 0.674 & 0.362 & 5.407 & 29.552 & \cellcolor{ourblue}19.036 & 0.619 & 0.449 & 7.166 & 21.753 & 21.396 & 0.718 & 0.320 & 5.451 & 35.970 & 26.13M \\
        FocalNet~\citep{focal2023} & ICCV'23 & 19.244 & \cellcolor{ourblue}0.679 & 0.367 & 5.933 & 37.687 & \cellcolor{ourpink}19.069 & \cellcolor{ourpink}0.628 & 0.435 & 5.967 & 36.892 & 21.422 & 0.717 & 0.320 & 5.675 & 38.538 & 3.74M \\
        ConvIR~\citep{cui2024revitalizing} & TPAMI'24 & \cellcolor{ourpink}19.432 & 0.678 & \cellcolor{ourpink}0.336 & 5.515 & 32.126 & 18.870 & 0.619 & \cellcolor{ourpink}0.426 & \cellcolor{ourpink}5.553 & 28.893 & 21.331 & 0.717 & 0.341 & 5.661 & 42.003 & 14.83M \\
        Fourmer~\citep{zhou2023fourmer} & ICML'24 & 19.367 & 0.675 & 0.359 & 5.506 & 30.973 & 18.872 & 0.619 & 0.489 & 6.212 & 27.266 & 21.422 & 0.716 & 0.330 & 5.653 & 36.194 & \cellcolor{ourblue}0.68M \\
        MambaIR~\citep{guo2025mambair} & ECCV'24 & 19.319 & \cellcolor{ourblue}0.679 & 0.355 & 6.033 & 43.266 & 18.872 & 0.622 & 0.449 & 6.331 & 34.889 & 21.426 & 0.722 & 0.325 & 6.066 & 44.614 & 4.31M \\
        Histoformer~\citep{sun2025restoring} & ECCV'24 & 17.666 & 0.439 & 0.481 & \cellcolor{ourpink}4.700 & \cellcolor{ourblue}12.315 & 16.573 & 0.346 & 0.607 & 6.729 & \cellcolor{ourpink}17.305 & 20.125 & 0.579 & 0.355 & \cellcolor{ourpink}4.336 & \cellcolor{ourpink}4.269 & 29.92M \\
        RAMiT~\citep{choi2024reciprocal} & CVPR'24 & 19.033 & 0.677 & 0.371 & 5.773 & 38.571 & 19.019 & 0.625 & 0.439 & 6.146 & 34.543 & 21.429 & 0.718 & 0.330 & 5.775 & 44.203 & \cellcolor{ourpink}0.30M \\
        AMIR~\citep{yang2024all} & MICCAI'24 & 19.118 & \cellcolor{ourpink}0.680 & 0.369 & 5.641 & 40.091 & 18.871 & \cellcolor{ourblue}0.627 & 0.440 & 6.075 & 32.879 & 21.292 & 0.717 & 0.346 & 5.441 & 38.380 & 23.54M \\
        AST~\citep{zhou2024AST} & CVPR'24 & 19.182 & 0.635 & 0.382 & 5.085 & \cellcolor{ourpink}9.012 & 17.046 & 0.606 & 0.448 & 6.182 & 24.973 & \cellcolor{ourblue}22.046 & \cellcolor{ourblue}0.731 & \cellcolor{ourblue}0.288 & 4.752 & 24.906 & 19.92M \\
        X-Restormer~\citep{chen2024xrestormer} & ECCV'24 & 18.034 & 0.659 & 0.372 & 5.201 & 36.810 & 18.603 & \cellcolor{ourpink}0.628 & 0.446 & 6.605 & 26.084 & \cellcolor{ourpink}22.317 & \cellcolor{ourpink}0.735 & 0.294 & 5.331 & 37.425 & 42.52M \\
        SFHformer~\citep{jiang2024fast} & ECCV'24 & 19.307 & 0.668 & \cellcolor{ourblue}0.351 & \cellcolor{ourblue}5.048 & 15.992 & 18.794 & 0.612 & \cellcolor{ourblue}0.432 & \cellcolor{ourblue}5.893 & \cellcolor{ourblue}18.287 & 21.231 & 0.696 & \cellcolor{ourpink}0.278 & \cellcolor{ourblue}4.636 & \cellcolor{ourblue}17.008 & 7.67M \\
        MambaIRv2~\citep{guo2024mambairv2} & CVPR'25 & \cellcolor{ourblue}19.391 & \cellcolor{ourpink}0.680 & 0.367 & 5.758 & 39.578 & 18.930 & 0.624 & 0.436 & 6.263 & 35.709 & 21.398 & 0.718 & 0.325 & 5.881 & 40.824 & 0.77M \\
        \bottomrule
    \end{tabular}
    \vspace{-5pt}
    \label{tab:results}
\end{table*}

\begin{figure*}[t]
\centering
\includegraphics[width=0.98\linewidth]{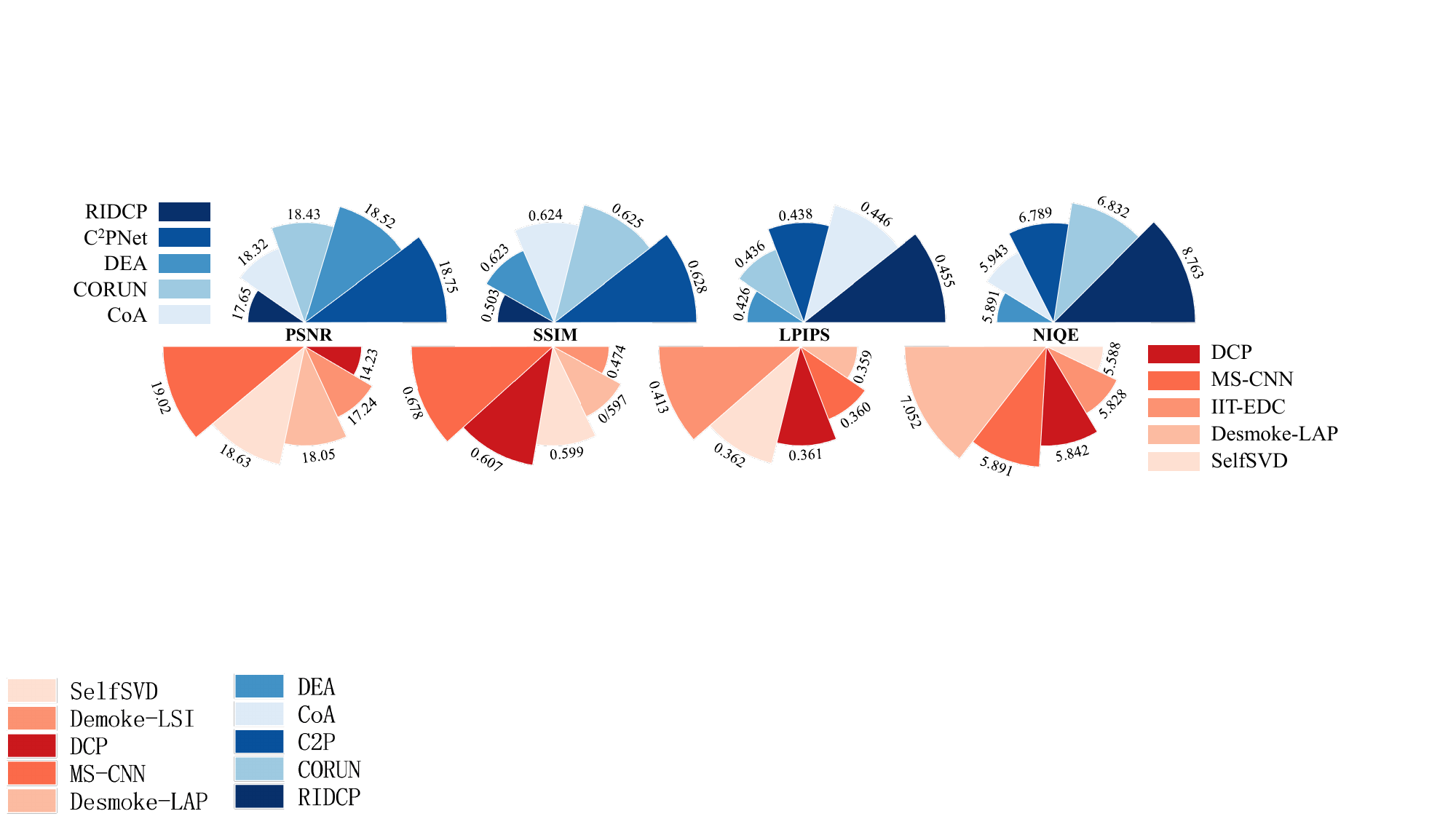}
\vspace{-0.7em}
\caption{Quantitative comparison for surgical image restoration with representative task-specific restoration models on the~\ourdata~test sets. The upper and lower panels comparatively demonstrate defogging and desmoking, respectively.}
\label{rose}
\vspace{-0.5em}
\end{figure*}

\begin{figure}[t!]
\centering
\includegraphics[width=\linewidth]{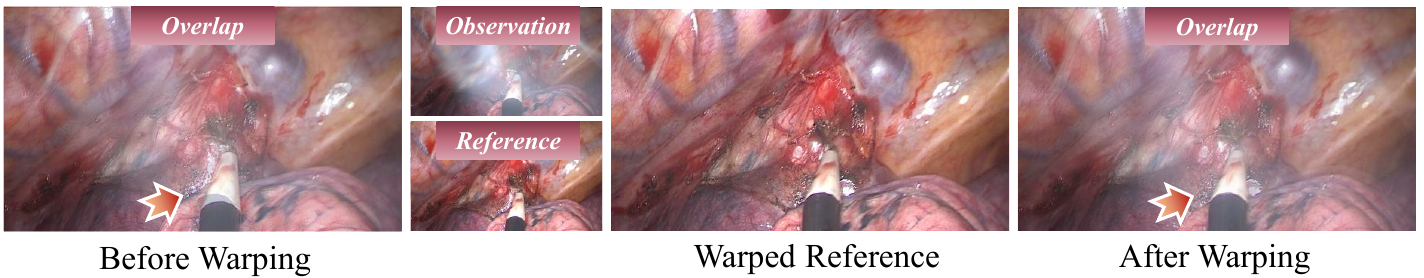}
\vspace{-1.5em}
\caption{The effect of our warping operations, which yields better alignment between paired GTs and input images.}
\label{warping}
\vspace{-10pt}
\end{figure}

\begin{figure*}[t]
\centering
\includegraphics[width=0.99\linewidth]{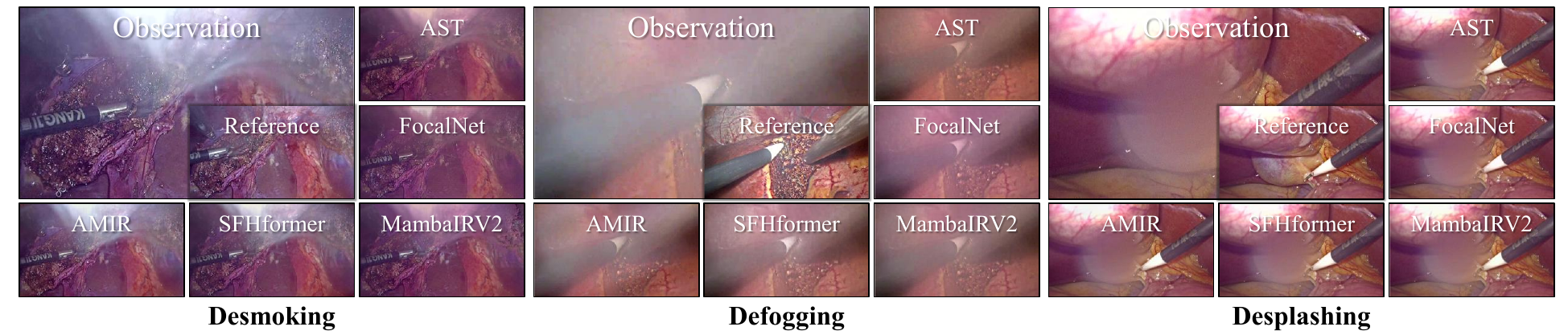}
\vspace{-0.5em}
\caption{Qualitative comparison for surgical image restoration on the~\ourdata~dataset. The methods selected for visual comparison were chosen from the top-five performing approaches in Table~\ref{tab:results}, based on their quantitative benchmark metrics.}
\label{comparison}
\vspace{-5pt}
\end{figure*}

\noindent\textbf{Evaluation Metrics.}
In line with general evaluation metrics for image restoration, we evaluate model performance using Peak Signal-to-Noise Ratio (PSNR), Structural Similarity Index (SSIM), and Learned Perceptual Image Patch Similarity (LPIPS) metrics. Referring to~\citep{wu2025self}, we also adopt non-reference metrics, \ie, Naturalness Image Quality Evaluator (NIQE) and Perceptual Index (PI), to evaluate the restoration results of three sub-degradation scenarios.

\noindent\textbf{Training Setting.}
As mentioned in~\secref{dataanalysis}, our dataset provides unaligned paired reference labels to facilitate the training process of restoration models.
To mitigate the impact of unaligned references, all comparison models use pre-trained PWC-Net~\citep{sun2018pwc} to estimate the optical flow of unaligned reference $\mathbf{UR}$ so that it aligns with the predicted restored image $\mathbf{P}$:
% $\mathbf{F}_{{UR} \rightarrow P} = \mathcal{O}(\mathbf{UR}, \mathbf{P})$,
\begin{equation}
\mathbf{F}_{{UR} \rightarrow P} = \mathcal{O}(\mathbf{UR}, \mathbf{P}),
\end{equation}
where $\mathcal{O}$ denotes the pre-trained optical flow network. Given the estimated optical flow $\mathbf{F}_{{UR} \rightarrow P}$, we apply the warping operation $\mathcal{W}$ to the $\mathbf{UR}$ image, transforming the unaligned label into aligned $\mathbf{UR}_{warp}$:
% $\mathbf{UR}_{warp} = \mathcal{W}(\mathbf{UR}, \mathbf{F}_{UR \rightarrow P}).$ 
\begin{equation}
\mathbf{UR}_{warp} = \mathcal{W}(\mathbf{UR}, \mathbf{F}_{UR \rightarrow P}).
\end{equation}
The visualization of the warping operations is shown in~\figref{warping}. Finally, $\mathbf{UR}_{warp}$ is spatially aligned with the restored image $\mathbf{P}$. The reconstruction loss can be expressed as
% $\mathcal{L}_{rec} = \sum_{i=1}^N||\mathbf{M}_{i} \odot ( \mathbf{UR}_{warp,i} - \mathbf{P}_i||_1.$
\begin{equation}
\label{eq:warpL1}
\mathcal{L}_{rec} = \sum_{i=1}^N||\mathbf{M}_{i} \odot ( \mathbf{UR}_{warp,i} - \mathbf{P}_i||_1.
\end{equation}
where $\odot$ is the pixel-wise multiplication, and Mi masks areas with imperfect optical flow to prevent trivial solutions. This allows the model to be trained and evaluated in the same way as the image restoration task with aligned GT. We apply this warping-based training and evaluation strategy to all models in our benchmark to ensure fairness.

\begin{figure*}[t]
\centering
\includegraphics[width=0.96\linewidth]{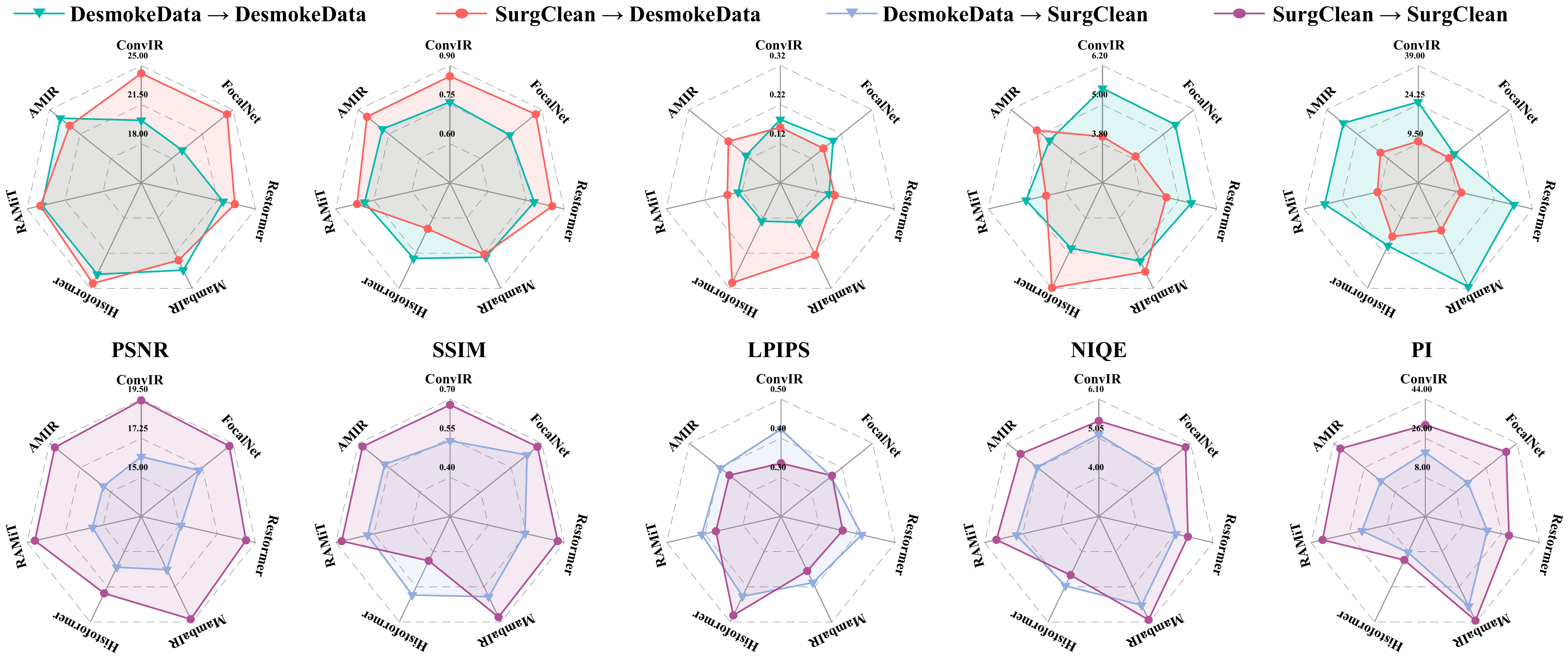}
\vspace{-0.3em}
\caption{Cross-dataset validation for desmoking between DesmokeData~\citep{xia2024new} and~\ourdata~datasets (Training dataset$\rightarrow$Testing dataset).}
\label{cross}
\vspace{-10pt}
\end{figure*}

\subsection{Benchmark Approaches}

We conduct a comprehensive evaluation of 22 representative methods on~\ourdata, including 12 generic and 10 task-specific models. Specifically, generic frameworks are built on different infrastructures, including CNN-based ConvIR~\citep{cui2024revitalizing} and FocalNet~\citep{focal2023}, transformer-based Restormer~\citep{zamir2022restormer}, X-Restormer~\citep{chen2024xrestormer}, AST~\citep{zhou2024AST}, RAMiT~\citep{choi2024reciprocal}, and Histoformer~\citep{sun2025restoring}, state space model-based MambaIR~\citep{guo2025mambair} and MambaIRv2~\citep{guo2024mambairv2}, and Fourier-based methods Fourmer~\citep{zhou2023fourmer} and SFHformer~\citep{jiang2024fast}.
Besides, we employ task-specific frameworks to validate the effectiveness in single degradation scenarios, including MS-CNN~\citep{wang2019multiscale}, Desmoke-LAP~\citep{pan2022desmoke}, IIT-EDC~\citep{salazar2020desmoking}, DCP~\citep{he2010single}, and SelfSVD~\citep{wu2025self} for desmoking, and C$^2$PNet~\citep{zheng2023curricular}, RIDCP~\citep{wu2023ridcp}, CORUN~\citep{fang2024real}, DEA~\citep{chen2024dea}, and CoA~\citep{ma2025coa} for defogging.

\subsection{Benchmark Results}
% 1. 自然场景weight直接推理
% 2. Fine tune

% desplashing 单独分析

\textbf{Quantitative Analysis.} 
We conduct a comprehensive comparison of representative generic and task-specific image restoration methods on \ourdata~for three tasks. 

As shown in~\tabref{tab:results}, existing generic methods still face unique challenges when dealing with complex degradation issues in real-world surgical environments. Specifically, ConvIR~\citep{cui2024revitalizing} achieves relatively superior PSNR and LPIPS scores, \ie, 19.432 and 0.336, for desmoking, but these results remain insufficient for high-quality reconstruction of smoke degradation during surgery. FocalNet~\citep{focal2023} achieves higher PSNR and SSIM values in defogging, but lower NIQE and PI values indicate that restored images still fall short of clean images in terms of authenticity and naturalness. For desplashing, X-Restormer~\citep{chen2024xrestormer} and AST~\citep{zhou2024AST} perform better in terms of PSNR and SSIM metrics, but the results require further improvement to meet the standards for clinical applications. In comparison, RAMiT~\citep{choi2024reciprocal} has only 0.3M parameters, which provides advantages when deployed on edge equipment. 

\begin{table*}[t]
\centering
\caption{Quantitative results on downstream tasks. All metrics were evaluated on the defogging samples of the SurgClean dataset.}
\vspace{-0.5em}
\label{tab:downstream}
\renewcommand{\arraystretch}{1.1}
\setlength{\tabcolsep}{3.0mm}
\scriptsize
\begin{tabular}{l | cccccc | cc | cc}
\toprule
\multirow{2}{*}{\textbf{Method}} &
\multicolumn{6}{c|}{\textbf{Scene Analysis}: Depth Estimation}  &
\multicolumn{4}{c}{\textbf{Scene Parsing}: Semantic (Left), Instrument (Right)} \\
\cline{2-11}
 & Abs Rel$\downarrow$ & Sq Rel$\downarrow$ & RMSE$\downarrow$ & $\delta_1$$\uparrow$ & $\delta_2$$\uparrow$ & $\delta_3$$\uparrow$
 & mIoU$\uparrow$ & mAcc$\uparrow$
 & mIoU$\uparrow$ & mAcc$\uparrow$ \\
\hline
Restormer~\citep{zamir2022restormer} &
0.3666 & 1.0348 & 2.3727 & 0.4855 & 0.7344 & 0.8569 &
0.4161 & 0.5869 & 0.7645 & 0.8799 \\

FocalNet~\citep{focal2023} &
0.3359 & 0.9202 & 2.2360 & 0.5234 & 0.7615 & 0.8740 &
0.4112 & 0.5830 & 0.7526 & 0.8716 \\

ConvIR~\citep{cui2024revitalizing} &
\cellcolor{ourpink}0.3302 & \cellcolor{ourpink}0.8885 & \cellcolor{ourpink}2.2030 & 0.5225 & \cellcolor{ourpink}0.7680 & \cellcolor{ourpink}0.8780 &
\cellcolor{ourblue}0.4194 & \cellcolor{ourblue}0.5877 & 0.7616 & 0.8797 \\

Fourmer~\citep{zhou2023fourmer} &
0.3536 & 0.9653 & 2.3477 & 0.5024 & 0.7436 & 0.8628 &
0.3929 & 0.5666 & 0.7593 & 0.8748 \\

Histoformer~\citep{sun2025restoring} &
0.4136 & 1.2328 & 2.5436 & 0.4362 & 0.6935 & 0.8294 &
0.2955 & 0.5284 & 0.7334 & 0.8817 \\

RAMiT~\citep{choi2024reciprocal} &
\cellcolor{ourblue}0.3326 & \cellcolor{ourblue}0.8978 & \cellcolor{ourblue}2.2273 & \cellcolor{ourpink}0.5290 & \cellcolor{ourblue}0.7624 & \cellcolor{ourblue}0.8764 &
0.4074 & 0.5790 & \cellcolor{ourblue}0.7646 & 0.8792 \\

AMIR~\citep{yang2024all} &
0.3428 & 0.9370 & 2.2745 & 0.5167 & 0.7527 & 0.8716 &
0.3811 & 0.5652 & \cellcolor{ourpink}0.7690 & \cellcolor{ourblue}0.8844 \\

AST~\citep{zhou2024AST} &
0.4096 & 1.2583 & 2.5460 & 0.4368 & 0.6894 & 0.8285 &
0.2600 & 0.5450 & 0.7400 & \cellcolor{ourpink}0.8950 \\

MambaIR~\citep{guo2025mambair} &
0.3400 & 0.9127 & 2.2698 & 0.5103 & 0.7503 & 0.8684 &
0.3876 & 0.5651 & 0.7642 & 0.8803 \\

X\text{-}Restormer~\citep{chen2024xrestormer} &
0.3338 & 0.9876 & 2.2650 & 0.5110 & 0.7502 & 0.8701 &
0.3350 & 0.5250 & 0.7550 & 0.8840 \\

SFHformer~\citep{jiang2024fast} &
0.3400 & 0.9158 & 2.2550 & 0.5137 & 0.7535 & 0.8688 &
0.3915 & 0.5681 & 0.7573 & 0.8737 \\

MambaIRv2~\citep{guo2024mambairv2} &
0.3348 & 0.9103 & 2.2384 & \cellcolor{ourblue}0.5244 & 0.7611 & 0.8723 &
\cellcolor{ourpink}0.4204 & \cellcolor{ourpink}0.5902 & 0.7551 & 0.8745 \\
\bottomrule
\end{tabular}
\vspace{-5pt}
\end{table*}

\begin{figure*}[t]
\centering
\includegraphics[width=0.94\linewidth]{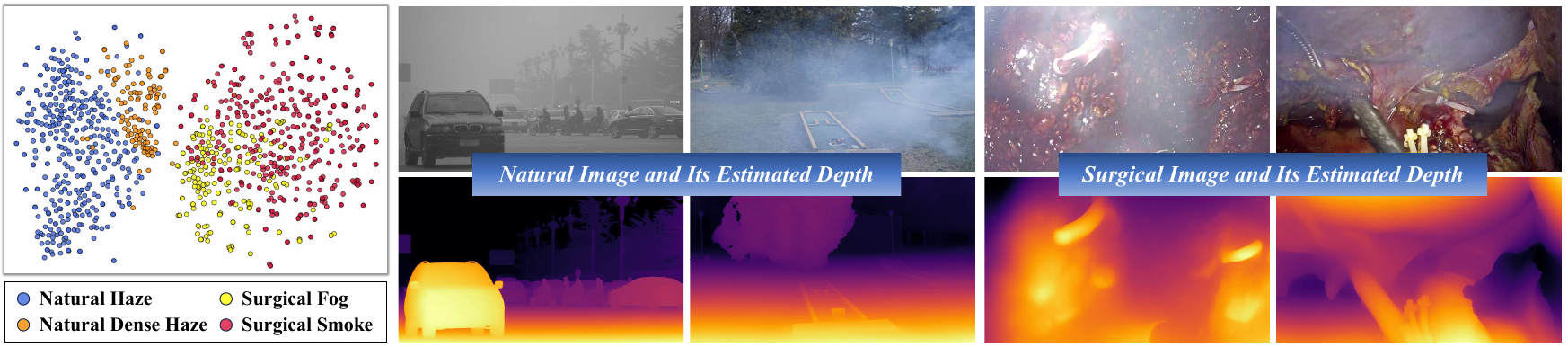}
\vspace{-0.7em}
\caption{Scene analysis between natural and surgical environments. The t-SNE visualization (Left) and depth examples (Middle/Right).}
\label{data_depth}
\vspace{-8pt}
\end{figure*}

\begin{figure*}[t]
\centering
\includegraphics[width=0.99\linewidth]{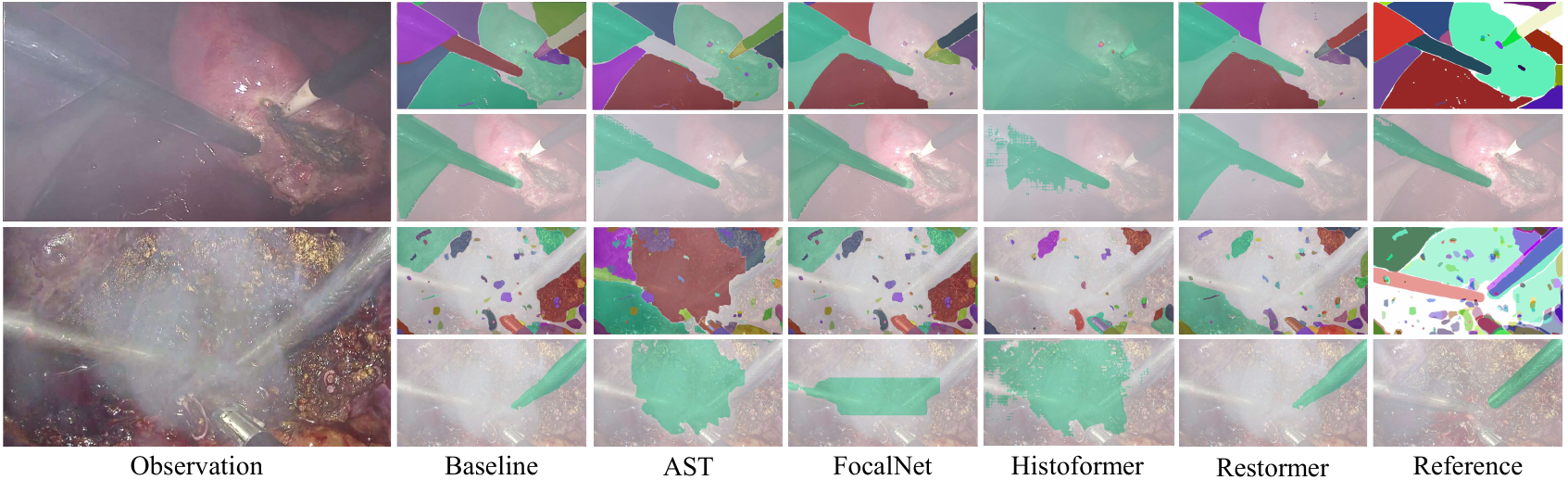}
\vspace{-0.5em}
\caption{Surgical scene parsing: SAM~\citep{kirillov2023segment} (Top) and MedSAM~\citep{ma2024segment} (Bottom).}
\label{data_seg}
\vspace{-5pt}
\end{figure*}

\figref{rose} illustrates the quantitative performance of task-specific restoration models for desmoking and defogging on the~\ourdata~test sets. 
% We can see that approaches specialized for desmoking achieve higher upper bounds on the PSNR metric than generic image restoration models, such as MS-CNN~\citep{wang2019multiscale} with a score of \pjl{19.726 compared to 18.630} achieved by AST~\citep{zhou2024AST}, but the overall performance is not significantly different. 
We can see that approaches specialized for desmoking do not yield better restoration performance compared to generic image restoration models, such as MS-CNN~\citep{wang2019multiscale} with a score of 19.02 compared to 19.43 achieved by ConvIR~\citep{cui2024revitalizing}.
For the defogging task, task-specific models also perform slightly worse than generic restoration models, highlighting the uniqueness and complexity of fog degradation in endoscopic scenes.
Overall, although various restoration techniques have developed rapidly in recent years, the specificity of degradation in multiple types of surgical scenarios, \eg, structural diversity and local damage, hinders the generalization of existing restoration models to real surgical environments.

% More visualization results can be found in the Appendix.

\noindent\textbf{Qualitative Analysis.}
% Due to space limitations, we conduct a qualitative comparison of representative methods for the desmoking task in~\figref{vis_res}\footnote{The detailed experimental comparisons and ablation studies can be found in the Appendix.}. 
% We provide visual analysis of several restoration methods with favorable quantitative results in typical desmoking, defogging, and desplashing scenarios.
As shown in~\figref{comparison}, most methods exhibit limited effectiveness in removing dense smoke within localized operating areas, and residual smoke still obscures anatomical structures. FocalNet~\citep{focal2023} and AST~\citep{zhou2024AST} perform better on tissue visibility, but still struggle to recover fine anatomical details.
For defogging, restoration models fail to completely suppress fog to provide a clear view, and structural distortion remains in the operating area.
Similarly, most approaches are ineffective in eliminating occlusions caused by local splashes.
In general, all benchmarking methods show limited performance when addressing various types of degradation in surgical scenarios, highlighting the uniqueness and persistent ESIR challenges.
% More visualization results can be found in \emph{Supp. D.2}.

\subsection{Cross-dataset Validation}

To further evaluate the usability and differences between our dataset and other open-source surgical image restoration datasets, we also invite the publicly available DesmokeData dataset ~\citep{xia2024new} with paired labels for cross-validation on the desmoking task.
DesmokeData contains a total of 961 degraded images along with corresponding paired labels. 
We randomly divided 761 images into the training set and 200 images for testing.
As shown in~\figref{cross}, we selected representative restoration models and conducted four sets of cross-validation experiments.
Comparing the results of (DesmokeData$\rightarrow$DesmokeData) and (\ourdata$\rightarrow$\ourdata), we can observe that almost all restoration methods perform better on DesmokeData than on~\ourdata~dataset. 
% This indicates that the~\ourdata~dataset is significantly higher than DesmokeData in terms of degradation complexity and restoration difficulty.
This indicates that the~\ourdata~dataset exhibits significantly higher degradation complexity and restoration difficulty compared to DesmokeData.
The reason probably lies in the presence of multi-level dense smoke occlusion and additional dynamic interference factors within~\ourdata.
When restoration models are trained on DesmokeData and then tested on~\ourdata, most methods show a significant decline in performance compared to those trained on in-domain data. In contrast, in experiment (\ourdata$\rightarrow$DesmokeData), the models trained on~\ourdata~perform more robustly when tested on DesmokeData.
This experiment demonstrates that models trained on~\ourdata~have stronger generalization performance, further highlighting the potential of our dataset in advancing the development of robust models.

\section{Beyond Pixel-Level Recovery}
% In this section, we attempt to look beyond the ``pixel-level recovery'' perspective from two aspects for surgical restoration, including scene analysis and scene parsing.

% \subsection{Downstream Tasks}

\subsection{Scene Analysis}
% Section~\ref{benchmark} explores a direct adaptation strategy by leveraging existing natural image restoration models for surgical scenarios, but extensive experiments reveal fundamental incompatibilities. 
Table~\ref{tab:downstream} (left) summarizes depth estimation results. Although FocalNet~\citep{focal2023} and Histoformer~\citep{sun2025restoring} achieve top defogging scores in Table~\ref{tab:results}, they fail to consistently lead in depth estimation. 
% Histoformer~\citep{sun2025restoring} shows a similar pattern.
Therefore, strong restoration performance does not guarantee superior depth estimation quality.
To explain this mismatch, we examine the underlying domain shift. As shown in Fig.~\ref{data_depth} (left), t-SNE reveals a clear distribution gap between surgical scenes and natural haze datasets~\citep{li2018benchmarking,ancuti2019ntire,ancuti2020nh}. Using a depth estimator~\citep{yang2024depth}, our scattering analysis in Fig.~\ref{data_depth} (right) further shows that endoscopic fog exhibits localized depth changes with abrupt transitions, whereas natural scenes display gradual, large-scale depth variations. These differences at both the statistical and imaging levels clarify why methods designed for natural image transfer poorly to surgical imagery.

\subsection{Scene Parsing}
We evaluated scene parsing with SAM for scene segmentation and MedSAM for instrument localization across surgical scenarios. As shown in Table~\ref{tab:downstream} (right), MambaIRv2 achieves better performance on two of four parsing metrics, yet its restoration scores in Table~\ref{tab:results} are relatively unremarkable, suggesting that higher restoration scores do not reliably yield stronger parsing.
Consistent with this observation, \figref{data_seg} shows that although restoration enhances visual quality, segmentation improvements remain unstable, particularly in structurally complex scenes. Balancing visual fidelity and task performance is thus crucial for clinical utility, motivating semantic-aware restoration frameworks.

% 为什么不做低光、超分等任务？
%数据量有限的客观因素
% 在实际情况下，烟雾、雾气和飞溅是否有可能同时发生？
% 把场景clean了之后会带来哪些临床价值，对于医生和患者的好处？

\section{Discussion}
The core objective of this study is to construct a multi-type endoscopic surgical image restoration (ESIR) benchmark for complex surgical scenarios. By establishing a standardized evaluation platform, we aim to advance restoration algorithms toward clinical applicability. Below are further analyses and insights into the~\ourdata~benchmark:

\emph{1) Suboptimal benchmark performance.} Both quantitative and qualitative analyses indicate that existing restoration methods are inadequate for handling the highly complex and variable degradations in surgical environments. Compared to natural image restoration, ESIR poses unique challenges, \eg, artifact blending with the background, localized occlusions, and the entanglement of diseased tissues with degraded areas, which significantly increase the complexity of surgical scenes. We attempted to explore the differences between surgical and natural images from the perspectives of structural depth and semantic understanding to provide insights for future research.

\emph{2) Selection of degradation types and data scale.} We selected three degradation scenarios (\ie, smoke, fog, and splashing) from 414 real surgical cases that experienced surgeons identified as directly affecting surgical efficiency. The distribution of samples across different degradation types reflects their relative frequency in the full dataset. We plan to continuously expand data with more real-world samples to facilitate further research in the ESIR community.

\emph{3) Simultaneous multi-type degradation.} While it is theoretically possible for multiple degradation types to occur simultaneously in the same scenario, such combined degradations were rarely observed in our data collection process. This may be due to the differing phases and conditions under which the three types of degradation occur. Thus, the first multi-type surgical restoration benchmark only focuses on samples with a single degradation type in a single scene. Future work will explore simultaneous degradations to enhance the ESIR diversity and challenge.

\emph{4) Clinical significance.} The surgical image restoration task provides surgeons with a clear surgical field of view, thereby significantly reducing operation time, minimizing surgical errors, and improving procedural efficiency. By eliminating unnecessary visual obstructions, surgeons can maintain focus during operations. Beyond enhancing image quality, ESIR also involves reconstructing complex anatomical details, which is critical for precise surgical navigation and shortening postoperative recovery times for patients.

\emph{5) Future research prospects.} Given the unique challenges of ESIR, future research can generate large amounts of synthetic data by simulating the imaging principles of surgical degradation to further reduce training costs. Besides, distilling semantic knowledge from large foundation models could improve the identification and localization of degraded regions. Finally, developing lightweight networks with optimized computational efficiency ensures practical deployment in clinical surgery.

\section{Conclusion}

We introduce~\ourdata, the first real-world, multi-type ESIR benchmark for real surgical degradations. 
The dataset contains 3,113 endoscopic images covering desmoking, defogging, and desplashing, with fine-grained severity and category labels. 
We provide a standardized evaluation protocol and comprehensive benchmark validation, illustrating that current restoration techniques still fall short of achieving clinically acceptable clarity. Beyond pixel metrics, we analyze structural differences between surgical and natural degradations and the impact on downstream surgical scene understanding. We expect SurgClean to catalyze research toward robust, clinically reliable intraoperative vision.

% We propose a comprehensive real-world laparoscopic surgical image restoration (LSIR) benchmark to address clinical demands in mitigating various scene degradation distortions. 
% We introduce~\ourdata, the first multi-type LSIR dataset tailored for real surgical scenarios. \ourdata~provides 1,020 images for desmoking, defogging, and desplashing, as well as fine-grained divisions based on degradation levels and categories. 
% We establish a standardized evaluation protocol and conduct systematic benchmark testing. The results illustrate that existing restoration techniques still fall short of meeting clinical requirements for a clear visual field. Beyond pixel-level perspective, we explore the structural differences and unique characteristics between surgical and natural scene degradations, and analyze the broad influence of restoration quality on downstream surgical scene understanding tasks.
% Through the proposed~\ourdata~benchmark, we aim to foster the emergence of future research to advance surgical restoration technology and ultimately contribute to reliable intraoperative visual assistance to reduce surgical risks.

\section*{Acknowledgements}

This work was supported in part by the Research Grants Council of the Hong Kong Special Administrative Region, China, under Project T45-401/22-N, in part by the Innovative Research Group Project of Hubei Province under Grants 2024AFA017, and in part by the National Natural Science Foundation of China (No. 62506060).

{
    \small
    \bibliographystyle{ieeenat_fullname}
    \bibliography{main}
}


\end{document}